\documentclass{ieeeaccess}
\usepackage{amsmath,amssymb,amsfonts}
\usepackage{graphicx}
\usepackage{textcomp}

\graphicspath{ {./figures/} }
\usepackage[pagebackref,breaklinks,bookmarks=false,colorlinks,linkcolor=black,anchorcolor=black,citecolor=black,filecolor=black,menucolor=black,runcolor=black,urlcolor=black]{hyperref}
\usepackage{amsthm}
\usepackage[numbers]{natbib}
\theoremstyle{definition}
\newtheorem{definition}{Definition}[]
\newtheorem{theorem}{Theorem}
\usepackage[ruled,vlined, linesnumbered]{algorithm2e}
\usepackage{booktabs}

\newcommand{\orcidicon}[1]{\href{https://orcid.org/#1}{\includegraphics[height=\fontcharht\font`\B]{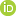}}}

\begin{document}
\history{Date of current version March 2, 2022.}
\history{Received March 2, 2022.}
\doi{ }

\title{The Recurrent Reinforcement Learning Crypto Agent}
\author{\uppercase{Gabriel Borrageiro}\orcidicon{0000-0002-0063-7103}, 
\uppercase{Nick Firoozye\orcidicon{0000-0002-6460-0406} and Paolo Barucca\orcidicon{0000-0003-4588-667X}}}
\address{Department of Computer Science, University College London, Gower Street, London, WC1E 6BT}

\markboth
{Borrageiro et al.: The Recurrent Reinforcement Learning Crypto Agent}
{Borrageiro et al.: The Recurrent Reinforcement Learning Crypto Agent}

\corresp{Corresponding author: Gabriel Borrageiro (e-mail: gabriel.borrageiro.20@ucl.ac.uk).}

\begin{abstract}
We demonstrate a novel application of online transfer learning for a digital assets trading agent. This agent uses a powerful feature space representation in the form of an echo state network, the output of which is made available to a direct, recurrent reinforcement learning agent. The agent learns to trade the XBTUSD (Bitcoin versus US Dollars) perpetual swap derivatives contract on BitMEX on an intraday basis. By learning from the multiple sources of impact on the quadratic risk-adjusted utility that it seeks to maximise, the agent avoids excessive over-trading, captures a funding profit, and can predict the market's direction. Overall, our crypto agent realises a total return of 350\%, net of transaction costs, over roughly five years, 71\% of which is down to funding profit. The annualised information ratio that it achieves is 1.46.
\end{abstract}

\begin{keywords}
online learning, transfer learning, echo state networks, recurrent reinforcement learning, financial time series
\end{keywords}

\titlepgskip=-15pt

\maketitle

\section{Introduction} \label{sec:intro}
Financial time series provide many modelling challenges for both researchers and practitioners. In some circumstances, data availability is sparse, and the datasets are vast in other circumstances; this impacts the choice of model and the learning style. In addition, financial time series are typically both autocorrelated and nonstationary, requiring approaches such as integer or fractional differencing \citep{ShumwayRobertH2011Tsaa} to remove these effects and facilitate correct feature selection; this is previously identified by \citeauthor{GrangerNewbold1974} \citep{GrangerNewbold1974} as leading to spurious regressions if not mitigated. Another approach to coping with nonstationarity is to allow models to learn continuously.

Against this backdrop, we extend our earlier work \citep{Borrageiro_RL_FX} where we combine sequential learning with transfer learning \citep{yang_transfer_learning_2020} and reinforcement learning \citep{SuttonBarto2018}. More concretely, we novelly transfer the learning of an echo state network \citep{Jaeger2002AdaptiveNS} to a direct, recurrent reinforcement learning agent \citep{Moody1998PerformanceFA} who must learn to trade digital asset futures, specifically the XBTUSD (Bitcoin versus US Dollar) perpetual swap on the BitMEX exchange. Our transfer learner benefits from an ample, dynamic reservoir feature space and can identify and learn from the different sources of impact on profit and loss, including execution costs, exchange fees, funding costs and price moves in the market.

Perhaps the main benefit of this paper will be for financial industry practitioners. In the numerous papers we researched on machine learning applications to financial trading, the researchers' emphasis tends to be on the novelty of the machine learning model, which inevitably has a high learning capacity. In practically all cases, good risk-adjusted returns are claimed, yet when one digs into the results in more detail, one invariably finds that trading costs are not fully accounted for. Such papers usually assume the execution of trades on the closing prices of sub-sampled data with certainty. However, only a price taker can obtain certainty of fill by crossing the bid/ask spread. The price taker then observes an immediate loss equal to the execution time half bid/ask spread. We barely see this cost accounted for, and even when it is, such as with the seminal work of \citeauthor{Moody1998PerformanceFA} \citep{Moody1998PerformanceFA}, a fixed execution cost is assumed. This fixed execution cost is never observed in reality; see, for example, figure 4 of \citeauthor{Borrageiro_RL_FX} \citep{Borrageiro_RL_FX}, which shows that the bid/ask spread varies by time of day. For many assets, especially in traditional finance, the bid/ask spread varies by day of the week as well; \citeauthor{Dacorogna2001AnIT} \citep{Dacorogna2001AnIT} demonstrate and discuss various examples of such stylised facts. Another approach commonly taken by academic financial trading researchers is to use supervised learning and sub-sample the high-frequency data into monthly time series. The main reason for doing this is to ameliorate excessive trading and thus high execution cost. Inevitably, we see the caveat emptor of a lack of experimental data and its impact on the generalisation performance of their chosen model. Furthermore, the down-sampling of data into lower frequencies typically makes bearable the slow training times of many models, such as deep q-learning networks.

We complete this section with a summary of the main contributions of this paper. Our meta-model can process data as a stream and learn sequentially; this helps it cope with the nonstationarity of the high-frequency order book and trading data. Furthermore, by using the vast high-frequency data, our model, which has a high learning capacity, avoids the kind of overfitting on a lack of data points that occurs with down-sampled data. We escape the problem of over-trading that is typically seen with the supervised learning model by learning the sensitivity of the change in risk-adjusted returns to the model's parameterisation. Stated another way, our model learns from the multiple sources of impact on profit and loss and targets the appropriate risk position. Finally, the scientific experiment that we conduct is representative of the conditions that would be observed in live trading; thus, we are confident that the resulting performance can realistically be transferred to industry use.

\section{Preliminaries}\label{sec:prelim}
This section provides a brief overview of the related ideas that we use in our experimentation, namely transfer learning, reservoir computing with echo state networks, and a form of policy gradients reinforcement learning. Finally, we conclude the section with a literature review of recent publications that apply reinforcement learning in cryptocurrency trading.

\subsection{Transfer Learning}\label{sec:transfer_learning}
Transfer learning refers to the machine learning paradigm in which an algorithm extracts knowledge from one or more application scenarios to help boost the learning performance in a target scenario \citep{yang_transfer_learning_2020}. Typically, traditional machine learning requires large amounts of training data. Transfer learning copes better with data sparsity by looking at related learning domains where data is sufficient. Even in a big data scenario such as streaming high-frequency data, transfer learning can benefit from learning immediately, where data is initially sparse, and a learner must begin providing forecasts when asked to do so. An increasing number papers focus on online transfer learning \citep{Zhao_OTL_2014,Bruno_OTL_2019, Wang_OTL_2020}. Following \citeauthor{Pan2010ASO} \citep{Pan2010ASO}, we define transfer learning as: 
\begin{definition}[transfer learning]
Given a source domain $\mathcal{D}_S$ and learning task $\mathcal{T}_S$, a target domain $\mathcal{D}_T$ and learning task $\mathcal{T}_T$, transfer learning aims to help improve the learning of the target predictive function $f_T(.)$ in $\mathcal{D}_T$ using the knowledge in $\mathcal{D}_S$ and $\mathcal{T}_S$, where $\mathcal{D}_S \neq \mathcal{D}_T$, or $\mathcal{T}_S \neq \mathcal{T}_T$.
\end{definition}

\subsection{Echo State Networks} \label{sec:esn}
Echo state networks are a form of recurrent neural network. They consist of a large, fixed, recurrent reservoir network, from which the desired output is obtained by training suitable output connection weights. Determination of the optimal output weights is solvable analytically, for example, sequentially with recursive least squares \citep{Jaeger2002AdaptiveNS}. Echo state networks are an example of the reservoir computing paradigm of understanding and training recurrent neural networks, based on treating the recurrent part (the reservoir) differently than the readouts from it \citep{Lukosevicius2012ReservoirCT}. Following \citeauthor{Jaeger2002AdaptiveNS} \citep{Jaeger2002AdaptiveNS}, the echo state property states that: 
\begin{definition}[echo state property]
If a network is started from two arbitrary states $\mathbf{x}_0$, $\tilde{\mathbf{x}}_0$ and is run with the same input sequence in both cases, the resulting state sequences $\mathbf{x}_T$, $\tilde{\mathbf{x}}_T$ converge to each other. If this condition holds, the reservoir network state will asymptotically depend only on the input history, and the network is said to be an echo state network.
\end{definition}
The echo state property is guaranteed if the dynamic reservoir weight matrix $\mathbf{W}^{hidden}$ is scaled such that its spectral radius $\rho(\mathbf{W}^{hidden})$, that is, its largest absolute eigenvalue, satisfies $\rho(\mathbf{W}^{hidden}) < 1$. This ensures that $\mathbf{W}^{hidden}$ is contractive. The mathematically correct connection between the spectral radius and the echo state property is that the latter is violated if $\rho(\mathbf{W}^{hidden}) > 1$ in reservoirs using the $\tanh$ function as neuron nonlinearity and for zero input \citep{Lukosevicius2009ReservoirCA}.

\citeauthor{Murray2019LocalOL} \citep{Murray2019LocalOL} states that the criteria for plausible modelling on how the brain might perform challenging time-dependent computations are locality and causality. As the echo state network uses a fixed, dynamic reservoir of weights, whose update depends on local information of inputs and activations, with fixed and random feedback, the model offers a plausible model of biological function. These biological aspects are explored by \citeauthor{Maass2004OnTC} \citep{Maass2004OnTC} concerning liquid state machines and more generally with spiking neural networks in \citeauthor{Samadi2017DeepLW} \citep{Samadi2017DeepLW}.

Numerous articles demonstrate echo state networks within a reinforcement learning context. For example, \citeauthor{ANN_ICANN_2006} \citep{ANN_ICANN_2006} propose a novel method that uses echo state networks as function approximators in reinforcement learning. They emphasise that echo state networks are promising candidates for partially observable problems where information about the past may improve performance, such as with k-order Markov processes. Since echo state networks are effectively linear function approximators acting on the internal state representation built from the previous observations, Gordon's \citep{Gordon2000ReinforcementLW} results about linear function approximators can be transferred to the echo state networks architecture. Building on this, they provide proof of convergence to a bounded region for echo state network training in the case of k-order Markov decision processes.

\citeauthor{Shi2017EchoSN} \citep{Shi2017EchoSN} seek to model the optimal energy management of an office. Time series inputs such as the real-time electricity rate, renewable energy and energy demand are made available to an echo state network q-learning model, which determines the optimal charging/discharging/idle strategies for the battery in the office so that the total cost of electricity from the grid can be reduced. 

\citeauthor{CHEN2021} \citep{CHEN2021} develop a fault-tolerant adaptive tracking control method fused with an echo state network, driven by reinforcement learning for Euler-Lagrange systems subject to actuation faults. Specifically, the echo state network implements an associative search network, a control gain network and an adaptive critic network, resulting in enhanced learning capabilities and stronger robustness against external uncertainties or disturbances, thus better controlling performance.

\Figure[!t]()[width=0.99\columnwidth]{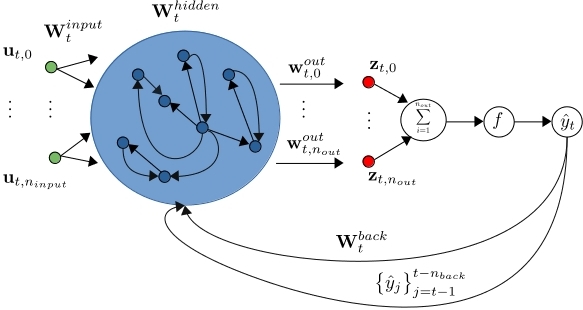}
 {A schematic of the echo state network (ESN), combined with the direct, recurrent reinforcement learning agent (DRL). The ESN part is inspired by the schematic of \citeauthor{Luko2012} \citep{Luko2012} and the DRL part is inspired by the McCulloch-Pitts schematic of \citeauthor{Bishop_1994} \citep{Bishop_1994}.\label{fig:esn_drl}}

\subsection{Policy Gradient Reinforcement Learning} \label{sec:pol_grad_rl}
In this section, we summarise our previous preliminary overview of the policy gradients method \citep{Borrageiro_RL_FX}. \citeauthor{Williams1992} \citep{Williams1992} introduced policy gradient methods in a reinforcement learning context. Whereas the majority of reinforcement learning algorithms tend to focus on action value estimation, learning the value of and selecting actions based on their estimated action values, policy gradient methods learn a parameterised policy that can select actions without the use of a value function. Williams also introduced his \textit{reinforce} algorithm
\begin{equation} \label{eq:policy_grad}
\Delta \mathbf{w}_{ij} = \eta_{ij}(r - b_{ij})\ln (\partial{\pi_i} / \partial{\mathbf{w}}_{ij}),
\end{equation}
where $\mathbf{w}_{ij}$ is the model weight going from the $j'th$ input to the $i'th$ output and $\mathbf{w}_i$ is the weight vector for the $i'th$ hidden processing unit in a network of such units, whose goal it is to adapt in such a way as to maximise the scalar reward $r$. The learning rate is $\eta_{ij}$ and the weight update of equation \ref{eq:policy_grad} is typically applied with gradient ascent. The reinforcement baseline $b_{ij}$, is conditionally independent of the model outputs $y_i$, given the network parameters $\mathbf{w}$ and inputs $\mathbf{x}_i$. The characteristic eligibility of $\mathbf{w}_{ij}$ is $\ln (\partial{\pi_i} / \partial{\mathbf{w}}_{ij})$, where $\pi_i(y_i = c, \mathbf{w}_i, \mathbf{x}_i)$ is the probability mass function determining the value of $y_i$ as a function of the parameters of the unit and its input. Baseline subtraction $r - b_{ij}$ plays an important role in reducing the variance of gradient estimators and \citeauthor{Sugiyama2015} \citep{Sugiyama2015} shows that the optimal baseline is given as
\[
b^* = \frac{\mathbb{E}_{p(r \lvert \mathbf{w})} \big [ r_t \| \sum_{t=1}^T \nabla \ln{\pi(a_t \lvert s_t, \mathbf{w}) \|^2} \big ]}{\mathbb{E}_{p(r \lvert \mathbf{w})} \big [\| \sum_{t=1}^T \nabla \ln{\pi(a_t \lvert s_t, \mathbf{w}) \|^2} \big ]},
\]
where the policy function $\pi(a_t \lvert s_t, \mathbf{w})$ denotes the probability of taking action $a_t$ at time t given state $s_t$, parameterised by $\mathbf{w}$. The main result of Williams's paper is 
\begin{theorem}
For any \textit{reinforce} algorithm, the inner product of $\mathbb{E}[\Delta \mathbf{w} \lvert \mathbf{w}]$ and $\nabla \mathbb{E}[r \lvert \mathbf{w}]$ is non-negative and if $\eta_{ij} > 0$, then this inner product is zero if and only if $\nabla \mathbb{E}[r \lvert \mathbf{w}] = 0$. If $\eta_{ij}$ is independent of $i$ and $j$, then $\mathbb{E}[\Delta \mathbf{w} \lvert \mathbf{w}] = \eta \nabla \mathbb{E}[r \lvert \mathbf{w}]$.
\end{theorem}
 This result relates $\nabla \mathbb{E}[r \lvert \mathbf{w}]$, the gradient in weight space of the performance measure $\mathbb{E}[r \lvert \mathbf{w}]$, to $\mathbb{E}[\Delta \mathbf{w} \lvert \mathbf{w}]$, the average update vector in weight space. Thus for any \textit{reinforce} algorithm, the average update vector in weight space lies in a direction for which this performance measure is increasing and the quantity $(r - b_{ij})\ln (\partial{\pi_i} / \partial{\mathbf{w}}_{ij})$ represents an unbiased estimate of $\partial{\mathbb{E}[r \lvert \mathbf{w}}] / \partial{\mathbf{w}_{ij}}$.

\subsubsection{Policy Gradient Methods in Financial Trading}
\citeauthor{Moody1998PerformanceFA} \citep{Moody1998PerformanceFA} propose to train trading systems and portfolios by optimising objective functions that directly measure trading and investment performance. Their model tries to target a position directly, and the model weights are adapted to maximise the performance measure. The performance function that they primarily consider is the differential Sharpe ratio. The annualised Sharpe ratio \citep{Sharpe1966} is
\[
  sr = 252^{0.5} \times \frac{\mu - r_f}{\sigma},
\]
where $\mu$ is the strategy's return, $\sigma$ is the standard deviation of returns, and $r_f$ is the risk-free rate. The differential Sharpe ratio is defined as
\[
  \frac{dsr_t}{d\tau} = \frac{b_{t-1} \Delta a_t - 0.5 a_{t-1} \Delta b_t}
  {(a_{t-1} - a_{t-1}^2)^{3/2}},
\]
where the quantities $a_t$ and $b_t$ are exponentially weighted estimates of the first and second moments of $r_t$
\[
  \begin{aligned}
    a_t &= a_{t-1} + \tau \Delta a_t = a_{t-1} + \tau (r_t - a_{t-1}) \\
    b_t &= b_{t-1} + \tau \Delta b_t = b_{t-1} + \tau (r_t^2 - b_{t-1}),
  \end{aligned}
\]
with $\tau \in (0, 1]$. They consider a batch gradient ascent update
\[
\Delta{\boldsymbol{\theta}_T = \eta \frac{dsr_T}{d\boldsymbol{\theta}}},
\]
where
\begin{equation}\label{eq:diff_sharpe}
\begin{split}
  \frac{dsr_T}{d\boldsymbol{\theta}} & = \frac{1}{T} \sum_{t=1}^T \frac{dsr_T}{dr_t} \Bigg\{ \frac{dr_t}{df_t} \frac{df_t}{d\boldsymbol{\theta}} + \frac{dr_t}{df_{t-1}} \frac{df_{t-1}}{d\boldsymbol{\theta}} \Bigg\} \\
  & = \frac{1}{T} \sum_{t=1}^T \Bigg\{ \frac{b_T - a_T r_t}{(b_T - a_T^2)^{3/2}} \Bigg\} \Bigg\{ \frac{dr_t}{df_t} \frac{df_t}{d\boldsymbol{\theta}} + \frac{dr_t}{df_{t-1}} \frac{df_{t-1}}{d\boldsymbol{\theta}} \Bigg\}.
\end{split}
\end{equation}
The reward 
\[
r_t = \Delta p_t f_{t-1} - \delta_t \lvert \Delta f_t \rvert
\]
depends on the change in reference price $p_t$, previous position $f_{t-1}$ and transaction costs $\delta_t$ which are applied only if there is a change in position $\lvert f_t - f_{t-1} \rvert > 0$. The position function is typically differentiable and bounded $-1 \leq f_t \leq 1$. This position function depends on the model inputs and parameters $f_t \triangleq f(\mathbf{x}_t; \boldsymbol{\theta}_t)$. The right half of equation \ref{eq:diff_sharpe} shows the dependency of the model parameters on the past sequence of trades. To correctly compute and optimise these total derivatives requires the use of recurrent algorithms such as backpropagation through time \citep{Rumelhart1986LearningIR, Werbos_bptt} or real-time recurrent learning \citep{RNN1989}. An undesirable property of the Sharpe ratio is that it penalises a model that produces returns larger than 
\[
  r^\ast_t = \frac{b_{t-1}}{a_{t-1}},
\]
which is counter-intuitive relative to most investors' notions of risk and reward \citep{Moody1998PerformanceFA}. \citeauthor{Gold2003} \citep{Gold2003} extends \citeauthor{Moody1998PerformanceFA} \citep{Moody1998PerformanceFA} work and investigates high-frequency currency trading with neural networks trained via recurrent reinforcement learning. He compares the performance of linear networks with neural networks containing a single hidden layer and examines the impact of shared system hyper-parameters on performance. In general, he concludes that the trading systems may be effective but that the performance varies widely for different currency markets, and simple statistics of the markets cannot explain this variability. 

\subsection{Reinforcement Learning Within Cryptocurrency Trading}
This section provides a brief literature review of reinforcement learning applied to cryptocurrency trading. Before performing this review, it is helpful to describe how returns net of transaction costs are generated when trading on an exchange. For example, assume that a single instrument is traded, such as Bitcoin versus the US Dollar. This instrument may be a cash or futures instrument. A gross profit (loss) is generated when Bitcoin is sold higher (lower) than its initial purchase price. Similar logic is applied for short positions, albeit with directions swapped. A net profit is observed by deducting various costs from the gross profit. These costs vary depending on the execution-style and are differentiated between price makers and price takers.

A price maker inserts quotes into an exchange limit order book and executes when a price taker removes the price maker's liquidity. The price maker captures half the bid/ask spread at the time of execution; the execution price is evaluated against the prevailing transaction mid-price ($0.5 \times [bid + ask]$), where $bid < mid < ask$. Bid/ask spreads in crypto tend to be as competitively priced as traditional financial instruments, although the exchange fees, usually a percentage of the notional traded, tend to be much higher. The price maker's fills are probabilistic, not inevitable. To compensate the market maker for the uncertainty incurred by providing liquidity and the risk of adverse selection, they capture half the bid/ask spread. The price taker obtains the certainty of immediate fill, subject to competing with other price takers for the same quoted liquidity. This certainty of fill comes at a cost, as the price taker must pay half the bid/ask spread and usually much higher exchange fees than price makers when trading crypto. 

A final cost that must be considered is funding. If one buys cash crypto without leverage, then it is plausible to consider no funding cost, and one can consider the purchase as self-funded. However, if one wants to sell cash crypto short, one needs to borrow the inventory; this attracts a funding cost. Furthermore, transacting in cryptocurrency futures on an exchange or contracts for difference in the over-the-counter market attracts a funding cost similar to what brokers charge for traditional financial instruments. Even if one trades futures directly on an exchange without a broker, some crypto derivatives contracts attract funding like traditional foreign exchange instruments do. For example, trading overnight foreign exchange exposes one to the interest rate differential between two currency pairs. Similarly, perpetual crypto swaps attract an intraday funding profit or loss, which ensures that the swap tracks the underlying cash instrument within tolerance. We now proceed with the literature review.

\citeauthor{crypto_portfolio_2017} \citep{crypto_portfolio_2017} supply a convolutional neural network with historical prices of a set of crypto assets as its input, outputting portfolio weights of the set of assets. The network is trained on less than a year of price data from the Poloniex cryptocurrency exchange. The training is done in a reinforcement learning manner, maximising the accumulated return as the reward function of the network. Using 30 minutely sampled closing prices, they conduct backtests which achieve ten-fold returns. In addition, they set exchange trading transaction fees of 25 basis points (25e-4) times the notional value traded of the base cryptocurrency. However, since they use closing prices to execute with a certainty of fill without applying half the observable bid/ask spread cost at the time of execution, the empirically observed backtest results do not represent the actual cost of trading and are thus more sanguine than reality.

\citeauthor{lee2018} \citep{lee2018} present a novel method to predict Bitcoin price movement using inverse reinforcement learning \citep{inverse_RL} and agent-based modelling. Their approach predicts the price by reproducing synthetic yet realistic behaviours of rational agents in a simulated market. Inverse reinforcement learning provides a systematic way to find the behavioural rules of each agent from Blockchain data by framing the trading behaviour estimation as a problem of recovering motivations from observed behaviour and generating rules consistent with these motivations. Once the rules are recovered, an agent-based model creates hypothetical interactions between the recovered behavioural rules, discovering equilibrium prices as emergent features through matching the supply and demand of Bitcoin. Their model is used to forecast the market's direction, and their results show that their proposed method can predict short-term market prices while outlining overall market trends. However, their experiments do not include the impact of holding risk, transaction or funding costs.

\citeauthor{deep_RL_crypto_2019} \citep{deep_RL_crypto_2019} apply deep reinforcement learning to trading Bitcoin. More precisely, double deep q-learning \citep{mnih2013playing} and duelling double deep q-learning \citep{dueling_net_2016} networks are trained in batch mode using 80\% of the four years of data that they have available. The remaining 20\% of the data is used for out of sample testing. Two reward functions are also tested: Sharpe ratio and profit reward functions. They find that the double deep q-learning trading system based on the Sharpe ratio reward function is the most profitable approach for trading Bitcoin. We note that the authors collect their data from Kaggle rather than from an actual crypto exchange and that they use minutely sampled open-high-low-close prices rather than actual order book bids and asks. As such, accurate transaction costs cannot be used in their experiment. Furthermore, no indication is made in their paper that they use any form of funding or exchange trading fees in their returns calculations.

\citeauthor{portfolio_deep_RL_2022} \citep{portfolio_deep_RL_2022} note that portfolio selection is difficult as the nonstationarity of financial time series and their complex correlations make the learning of feature representation challenging. They propose a cost-sensitive portfolio selection method with deep reinforcement learning. Specifically, a novel two-stream portfolio policy network is devised to extract price time series patterns and asset correlations, while a new cost-sensitive reward function is developed to maximise the accumulated return and constrain costs via reinforcement learning. They empirically evaluate their proposed method on real-world datasets from the Poloniex crypto exchange. Promising results demonstrate the effectiveness and superiority of the proposed method in terms of profitability, cost-sensitivity and representation abilities. Once more, however, transaction costs are not fully accounted for in their experiment. For example, they assume a fixed transaction cost of 25 basis points times the notional value traded of the base cryptocurrency; however, they also use open-high-low-close prices, sampled every 30 minutes. The closing prices they use assume that execution is immediate. In reality, however, immediate execution may only be achieved by crossing the spread; this additional cost must be modelled, which usually turns a theoretically profitable strategy that executes at the closing price or mid-price into a loss-making one.

\section{The Research Experiment}
We begin with a discussion of the research data we use in our experiment, followed by an elucidation of the research methods and a description of the experiment results. As a high-level summary, our experiment aims to explore transfer learning using a source model, an echo state network and a target model, a direct, recurrent, reinforcement learning agent. The objective of this meta-model is to learn to trade digital asset futures, specifically perpetual contracts on the BitMEX crypto exchange. Finally, the dynamical reservoir of the echo state network acts as a powerful nonlinear feature space; this is fed into the upstream recurrent reinforcement learner, who is aware of the various sources of impact on profit or loss and learns to target the desired position.

\subsection{The BitMEX XBTUSD Perpetual Swap}\label{sec:xbtusd}
\begin{figure}[ht]
 \centering
  \includegraphics[width=.99\columnwidth]{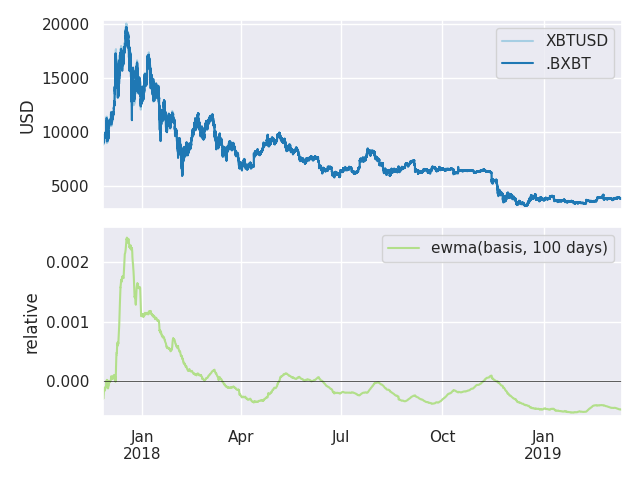}
  \caption{XBTUSD basis during the bear market of 2019}
  \label{fig:xbtusd_basis}
\end{figure}

The data that we experiment with is from the BitMEX cryptocurrency derivatives exchange. In 2016 they launched the XBTUSD perpetual swap, where clients trade Bitcoin against the US Dollar. The perpetual swap is similar to a traditional futures contract, except there is no expiry or settlement. It mimics a margin-based spot market and trades close to the underlying reference index price. A funding mechanism is used to tether the contract to its underlying spot price. In contrast, a futures contract may trade at a significantly different price due to the basis
\[
  basis_t = futures_t - index_t.
\]
The basis means different things in different markets. For example, in the oil market, the demand for spot oil can outpace the demand for futures oil, especially if OPEC withholds supply, leading to a higher spot price; this results in a futures curve in a state of backwardation. The crypto futures normally trade in a state of contango, where the futures prices trade at a higher rate than the spot prices. Backwardation or contango in crypto markets does not represent supply and demand shortages in an economic sense but rather reflects risk appetite for crypto. Similar effects happen in the equity markets. As with the equity markets, crypto market participants can take risks in the futures market more easily. The spot markets typically do not offer leverage, and the trader must have inventory in the exchange to trade. In contrast, futures allow traders to sell assets short with leverage and without borrowing the underlying asset. However, what is required is a margin or deposit to fund the position. Figure \ref{fig:xbtusd_basis} shows the basis in relative terms for the XBTUSD perpetual swap during the bear market of 2018. The mid-price of the perpetual swap is compared against the underlying index it tracks, .BXBT. The relative basis is computed as
\begin{equation} \label{eq:basis}
  rbasis_t = \frac{futures_t - index_t}{index_t}.
\end{equation}
Before this bear market, Bitcoin hit a then all-time high of \$20,000, and the 100-day exponentially weighted moving average of relative basis was very positive in late 2017. For most of 2018 and 2019, the basis was largely negative, reflecting the cash market sell-off from all-time highs to circa USD 3,000.

\subsubsection{Funding}
The funding rate for the perpetual swap comprises two parts: an interest rate differential component and the premium or discount of the basis. We denote this funding rate as $\kappa_t$. The interest differential reflects the borrowing cost of each currency involved in the pair
\[
  e_t = \frac{e^{quote}_t - e^{base}_t}{T}.
\]
For example, with the XBTUSD perpetual swap, the interest rate of borrowing in US Dollars is denoted by $e^{quote}_t$ and the interest rate of borrowing in Bitcoin is represented by $e^{base}_t$. As funding occurs every 8 hours, $T=8$. The basis premium/discount component is computed in a manner similar to equation \ref{eq:basis}, with some subtleties applied to minimise market manipulation, such as the use of time and volume weighted average prices. The funding rate is finally
\begin{equation} \label{eq:funding}
  \kappa_t = b_t + max(min[\zeta e_t - b_t],-\zeta),
\end{equation}
where $\zeta$ is a basis cap, typically 5 basis points (0.05\%).
When the basis is positive, traders with long positions (buy XBT, sell USD) will pay those with short positions (sell XBT, buy USD). Reciprocally, shorts pay longs when the basis is negative.

\subsection{The Recurrent Reinforcement Learning Crypto Agent}
We begin with a description of the dynamic reservoir feature space, the resultant learning of which is transferred to the direct, recurrent reinforcement learner, which targets the desired risk position. It is worth highlighting where our approach deviates from the traditional use of echo state networks within a reinforcement learning context. Figure \ref{fig:esn_drl} demonstrates visually that the target labels, the so-called teacher signal, can be fed back into the dynamic reservoir. Equally, one could apply a regression layer to the echo state network and feed the resulting forecasts into the dynamic reservoir. Both approaches make the echo state network recurrent then. We value this recurrent nature in a trading context, as we wish to feed information about the current position back into the model. However, rather than treating this exercise as a value function estimation task as with \citeauthor{ANN_ICANN_2006} \citep{ANN_ICANN_2006}, we feed the augmented, dynamic reservoir features of the echo state network to a direct recurrent reinforcement learner. By differentiating a quadratic utility function with respect to the recurrent reinforcement learner's parameters, with feedback connections from the agent's past positions fed back into the echo state network dynamic reservoir, the agent learns from the various sources of impact on profit and loss and targets the appropriate position that maximises the risk-adjusted reward.

\subsubsection{The Dynamic Reservoir Feature Space} \label{sec:dyn_res_feat_space}
Denote as $\mathbf{u}_t$, a vector of external inputs to the system, which is observed at time $t$. In the context of this experiment, such external input would include order book, transaction and funding information. These features may come from the instrument being traded, exogenous instruments, or both. Initialise the external input weight matrix $\mathbf{W}^{input} \in \mathbb{R}^{n_{hidden} \times n_{input}}$, where the weights are drawn at random; a draw from a standard normal would suffice. Here, $n_{hidden}$ denotes the number of hidden processing units in the internal dynamical reservoir and $n_{input}$ is the number of external inputs, including a bias term. Next we initialise the hidden processing units weight matrix, $\mathbf{W}^{hidden} \in \mathbb{R}^{n_{hidden} \times n_{hidden}}$. The procedure detailed by \citeauthor{Yildiz2012RevisitingTE} \citep{Yildiz2012RevisitingTE} is 
\begin{itemize}
\item Initialise a random matrix $\mathbf{W}^{hidden}$, all with non-negative entries.
\item Scale $\mathbf{W}^{hidden}$ such that its spectral radius $\rho(\mathbf{W}^{hidden}) < 1$.
\item Change the signs of a desired number of entries of $\mathbf{W}^{hidden}$ to get negative connection weights as well.
\item Sparsify $\mathbf{W}^{hidden}$ with probability $P(\alpha), 0 \ll \alpha < 1$, setting those elements to zero.
\end{itemize}
This procedure is guaranteed to ensure the echo state property for any input. Intuitively, a recurrent neural network has the echo state property concerning an input signal $\mathbf{u}_t$, if any initial network state is forgotten or washed out when the network is driven by $\mathbf{u}_t$ for long enough \citep{Jaeger2017UsingCT}. The model supports recurrent connections from either a teacher signal $\mathbf{y}_t \in \mathbb{R}^{n_{back}}$ or model output $\hat{\mathbf{y}}_t \in \mathbb{R}^{n_{back}}$. These are connected to the model via the weight matrix $\mathbf{W}^{back} \in \mathbb{R}^{n_{hidden} \times n_{back}}$, whose weights are initialised at random from a standard normal. Note that $\mathbf{W}^{input}$, $\mathbf{W}^{hidden}$ and $\mathbf{W}^{back}$, have weights that remain fixed.

Finally, initialise the output weight vector $\mathbf{w}^{out}_0 \in \mathbb{R}^{n_{input} + n_{hidden} + n_{back}}$. It is at this point that our procedure differs from the original echo state network formulation shown by \citeauthor{Jaeger2002AdaptiveNS} \citep{Jaeger2002AdaptiveNS}. There, $\mathbf{w}^{out}$ is a matrix $\mathbf{W}^{out} \in \mathbb{R}^{n_{back} \times (n_{input} + n_{hidden} + n_{back})}$ and the performance measure of the model is the quadratic loss
\[
  \min_{\mathbf{W}^{out}} \frac{1}{2T} \sum_{t=1}^T (y_t - \mathbf{W}^{out} \mathbf{z}_t)^2.
\]
The reasons will become apparent shortly when we detail the model's direct, recurrent reinforcement learning part. But first, we must describe how we create the augmented state of the system, $\mathbf{z}_t \in \mathbb{R}^{n_{input} + n_{hidden} + n_{back}}$. Firstly, we initialise a zero-valued internal state vector $\mathbf{x}_0 \in \mathbb{R}^{n_{hidden}}$. Then at time $t$, we compute the recurrent internal state
\[
  \mathbf{x}_t = f_{hidden}(\mathbf{W}^{input} \mathbf{u}_t + \mathbf{W}^{hidden} \mathbf{x}_{t-1} + \mathbf{W}^{back}_t \hat{\mathbf{y}}_t),
\]
where $f_{hidden}(.)$ is typically a squashing function such as the hyperbolic tangent. The augmented, recurrent system state is then
\begin{equation} \label{eq:augmented_state_vector}
  \mathbf{z}_t = [\mathbf{u}_t, \mathbf{x}_t, \hat{\mathbf{y}}_t].
\end{equation}
Equation \ref{eq:feedback_vector} defines what $\hat{\mathbf{y}}_t$ represents, namely the past desired positions of the direct, recurrent reinforcement learning agent.

\subsubsection{Direct Recurrent Reinforcement Learning} \label{sec:dir_rrl}
The augmented, internal feature state, $\mathbf{z}_t$, is now fed into the upstream model, a direct, recurrent neural network, whose performance measure is a quadratic utility function of reward and risk. For the reader's benefit and the fact that we use the same target transfer learning model, we describe the dynamics of the direct, recurrent reinforcement learner in a manner similar to our earlier work \citep{Borrageiro_RL_FX}. \citeauthor{Sharpe2007} \citep{Sharpe2007} discusses asset allocation as a function of expected utility maximisation, where the utility function may be more complex than that associated with mean-variance analysis. Denote the expected utility for a single asset portfolio as
\begin{equation} \label{eq:utility}
  \upsilon_t = \mu_t - \frac{\lambda}{2}\sigma_t^2,
\end{equation}
where the expected return $\mu_t$ and variance of returns $\sigma_t^2$ may be estimated in an online fashion with exponential decay
\begin{equation} \label{eq:expected_reward}
\mu_t = \tau \mu_{t-1} + (1 - \tau) r_t, 
\end{equation}
\[
\sigma_t^2 = \tau \sigma_{t-1}^2 + (1 - \tau) (r_t - \mu_t)^2.
\]
The risk appetite constant $\lambda > 0$, can be set as a function of an investor's desired risk-adjusted return, as demonstrated by \citeauthor{grinold2019advances} \citep{grinold2019advances}. Define the annualised information ratio as the risk-adjusted differential reward measure, where the difference is taken against a benchmark or baseline strategy
\[
ir_t = 252^{0.5} \times \frac{\mu_t - b_t}{\sigma_t}.
\]
Substituting the non-annualised information ratio into the quadratic utility and differentiating it against the risk, we obtain a suitable value for the risk appetite parameter
\[
  \lambda = \frac{ir_t}{\sigma_t}.
\]
The net returns whose expectation and variance we seek to learn, are decomposed as
\begin{equation} \label{pnl}
  r_t = \Delta p_t f_{t-1} - \delta_t \lvert \Delta f_t \rvert - \kappa_t f_t,  
\end{equation}
where $\Delta p_t$ is the change in reference price, typically a mid price
\[
\Delta p_t = 0.5 \times (bid_t + ask_t - bid_{t-1} - ask_{t-1}),
\]
$\delta_t$ represents the execution cost for a price taker
\begin{equation} \label{eq:price_taker}
\delta_t = 0.5 \times (ask_t - bid_t),
\end{equation}
$\kappa_t$ is the funding cost (see subsection \ref{sec:xbtusd}) and $f_t$ is the desired position learnt by the recurrent reinforcement learner
\begin{equation}\label{eq:position}
  f_t = tanh \big ( [\mathbf{w}^{out}_t]^T\mathbf{z}_t \big ).
\end{equation}
The model is maximally short when $f_t = -1$ and maximally long when $f_t = 1$. The past positions of the model are used as the feedback connections for equation \ref{eq:augmented_state_vector}
\begin{equation} \label{eq:feedback_vector}
    \hat{\mathbf{y}}_t = [f_{t-n_{back}}, ..., f_{t-1}].
\end{equation}

The goal of our recurrent reinforcement learner is to maximise the utility in equation \ref{eq:utility}, by targeting a position in equation \ref{eq:position}. To do this, we apply an online optimisation update of the form
\[
\mathbf{w}^{out}_t = \mathbf{w}^{out}_{t-1} + \nabla \upsilon_t \equiv \Delta \mathbf{w}^{out}_t + \frac{d\upsilon_t}{d\mathbf{w}^{out}_t},
\]
where the weight update procedure is an extended Kalman filter for neural networks \citep{Williams_rnn_ekf, Haykin2001}, albeit modified for reinforcement learning in this context; sequential updates are applied as per algorithm \ref{algo:ekf}.

\SetKwInput{KwRequire}{Require}
\SetKwInput{KwInit}{Initialise}  
\SetKwInput{KwInput}{Input}                
\SetKwInput{KwOutput}{Output}   
{
\centering
\vspace{0.25cm}
\begin{algorithm}
\SetAlgoLined
\KwRequire{$\beta$, $\tau$
\newline {\small \tcp{$\beta \geq 0$ is a Ridge penalty.}}
\tcp{$0 \ll \tau \leq 1$ is an exponential decay factor.}}
\KwInit{
$d = n_{input} + n_{hidden} + n_{back}$

$\mathbf{w}^{out}= \mathbf{0}_d$, $\mathbf{P} = \mathbf{I}_d / \beta$

{\small \tcp{$\mathbf{0}_d$ is a zero vector in $\mathbb{R}^{d}$.}}

{\small \tcp{$\mathbf{P}$ is the precision matrix in $\mathbb{R}^{d \times d}$.}}

}

\KwInput{$\nabla \upsilon_t$}
\KwOutput{$\mathbf{w}^{out}_t$}

$q = 1 + \nabla \upsilon_t^T \mathbf{P}_{t-1} \nabla \upsilon_t / \tau$

$\mathbf{k} = \mathbf{P}_{t-1} \nabla \upsilon_t / (q \tau)$

$\mathbf{w}^{out}_t = \mathbf{w}^{out}_{t-1} + \mathbf{k}$

$\mathbf{P}_t = \mathbf{P}_{t-1} / \tau - \mathbf{k} \mathbf{k}^T q$

$\mathbf{P}_t = \mathbf{P}_t \tau$ {\small \tcp{variance stabilisation}}

\caption{extended Kalman filter}\label{algo:ekf}
\end{algorithm}
\vspace{0.25cm}
}

Above, $\mathbf{P}_t$ is an approximation to $\nabla^2 \upsilon_t$, the inverse Hessian of the utility function $\upsilon_t$ with respect to the model weights $\mathbf{w}^{out}_t$. We decompose the gradient of the utility function with respect to the recurrent reinforcement learner's parameters as follows
\begin{equation} \label{eq:factoredDerivs}
\begin{split}
\nabla \upsilon_t & = \frac{d\upsilon_t}{dr_t} \bigg\{ \frac{dr_t}{df_t} \frac{df_t}{d\mathbf{w}^{out}_t} + \frac{dr_t}{df_{t-1}} \frac{df_{t-1}}{d\mathbf{w}^{out}_{t-1}} \bigg\} \\
 & = \Bigg\{  \frac{d\upsilon_t}{d\mu_t} \frac{d\mu_t}{dr_t} + \frac{d\upsilon_t}{d\sigma_t^2} \frac{d\sigma_t^2}{dr_t} \Bigg\} \Bigg\{ \frac{dr_t}{df_t} \bigg[ \frac{\partial f_t}{\partial \mathbf{w}^{out}_t} + \frac{\partial f_t}{\partial f_{t-1}} \frac{\partial f_{t-1}}{\partial \mathbf{w}^{out}_{t-1}} \bigg] \\
 & + \frac{dr_t}{df_{t-1}} \bigg[ \frac{\partial f_{t-1}}{\partial \mathbf{w}^{out}_{t-1}} + \frac{\partial f_{t-1}}{\partial f_{t-2}} \frac{\partial f_{t-2}}{\partial \mathbf{w}^{out}_{t-2}} \bigg] \Bigg\}.
\end{split}
\end{equation}

The constituent derivatives for the left half of equation \ref{eq:factoredDerivs} are:
\[
  \frac{d\upsilon_t}{dr_t} = (1-\eta)[1 - \lambda(r_t - \mu_t)],
\]
\[
  \frac{dr_t}{df_t} = -\delta_t \times sign(\Delta f_t) - \kappa_t,
\]
\[
\begin{split}
 \frac{df_t}{d\mathbf{w}^{out}_t} & = \mathbf{z}_t [1 - \tanh^2{([\mathbf{w}^{out}_t]^T \mathbf{z}_t)}] \\
 & + \mathbf{w}^{out}_{t,n}[1 - \tanh^2{([\mathbf{w}^{out}_t]^T \mathbf{z}_t)}] \\
 & \times \mathbf{z}_{t-1} [1 - \tanh^2{([\mathbf{w}^{out}_{t-1}]^T \mathbf{z}_{t-1})}],
\end{split}
\]
where $n = n_{input} + n_{hidden} + n_{back} - 1$, using $0$ as the starting index.

\subsection{Experiment Design}
We put a recurrent reinforcement learning crypto agent to work by trading the XBTUSD (Bitcoin vs US Dollar) perpetual swap on BitMEX. We transfer the output of the source model, the dynamic reservoir feature space of subsection \ref{sec:dyn_res_feat_space}, to the target model, the direct recurrent reinforcement learner of subsection \ref{sec:dir_rrl}, who learns to target a risk position directly. Finally, we use five minutely sampled intraday data. The choice of this sampling rate is driven by the throttle imposed by the vendor on retrieving historical data; if we could obtain the raw, asynchronously delivered tick data promptly, we would do so. Nevertheless, we are still using $365 \times 5 \times 1440 / 5 = 525600$ observations in our experiment. Our performance evaluation procedure involves the following:
\begin{itemize}
  \item Construct input features from the order book and trade information made available by  BitMEX for XBTUSD.
  \item Feed these input features into an echo state network, with $n_{hidden} = 100$, $n_{back} = 10$ and the percentage of reservoir units $\mathbf{W}^{hidden}$ that are sparsified, set to $\alpha = 0.75$.
  \item Feed the output of the echo state network (equation \ref{eq:augmented_state_vector}) into a direct, recurrent reinforcement learner (subsection \ref{sec:dir_rrl}). 
  \item Set the risk appetite constant $\lambda=0.00001$ for quadratic utility equation \ref{eq:utility}.
  \item Set the ridge penalty $\beta = 1$ and the exponential decay factor $\tau = 0.999$ for the extended Kalman filter of algorithm \ref{algo:ekf}.
  \item Backtest the entire history as a test set.
  \item Learn sequentially online to target the desired position.
  \item Force the agent to trade as a price taker, who incurs an execution cost equal to equation \ref{eq:price_taker} plus exchange fees, which are set to 5 basis points (0.05\%).
  \item For non-zero risk positions, apply the appropriate funding profit or loss as per equation \ref{eq:funding}.
  \item Monitor equation \ref{eq:expected_reward}, the expected net reward of the strategy. The crypto agent can trade freely if $\mu_t \ge 0$. Otherwise, close the position and wait for an opportunity to enter the market again.
\end{itemize}

\subsection{Results}
\begin{figure}[ht]
 \centering
  \includegraphics[width=.99\columnwidth]{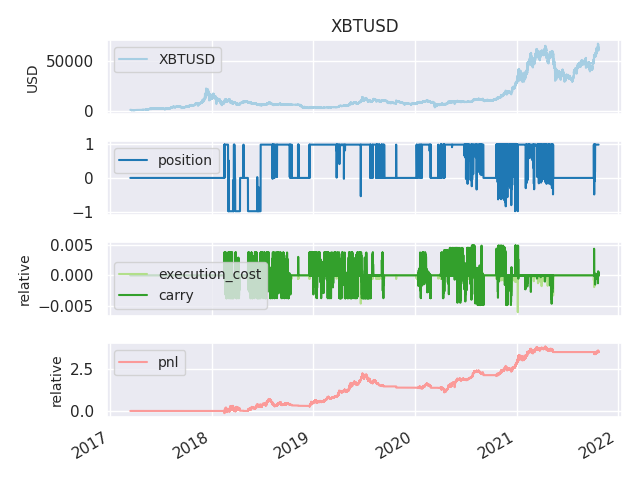}
  \caption{cumulative returns for the XBTUSD crypto agent}
  \label{fig:drl_esn_XBTUSD}
\end{figure}

Table \ref{tab:results} and figure \ref{fig:drl_esn_XBTUSD} show the results of the experiment. The crypto agent achieves a total return of just under 350\% over a test set that is less than five years. The associated annualised information ratio is 1.46. Denoting the maximally short position as -1, no position as 0 and the maximally long position as 1, we see that the agent averages a position of 0.41. Thus there is a bias toward the agent maintaining a long position, which is desirable, as Bitcoin has appreciated against the US Dollar over this period. We see visual evidence in figure \ref{fig:drl_esn_XBTUSD} that on occasion, the crypto agent abstains from trading, or rather is forced to take no position; this will happen during periods when the predictive performance of the agent decreases relative to execution and funding costs. Our crypto agent also captures a 71\% cumulative return due to earning funding, which is expected as the agent learns to target the appropriate position that maximises its quadratic utility, and funding is one of the drivers of this utility. The total execution cost and exchange fees that the agent pays out is -54\%.

\begin{table}[!htb]
\caption{daily pnl statistics for the XBTUSD crypto agent}
 \centering
 \begin{tabular}{lllll}
 \toprule
 {} & position & execution & carry & pnl \\
 \midrule
 count & 1684 & 1684 & 1684 & 1684 \\
 mean & 0.405 & -0.000 & 0.000 & 0.002 \\
 std & 0.579 & 0.001 & 0.005 & 0.027 \\
 min & -0.990 & -0.019 & -0.014 & -0.155 \\
 25 \% & 0.000 & 0.000 & -0.000 & 0.000 \\
 50 \% & 0.000 & 0.000 & 0.000 & 0.000 \\
 75 \% & 0.990 & 0.000 & 0.000 & 0.003 \\
 max & 0.990 & 0.000 & 0.014 & 0.177 \\
 sum & {} & -0.542 & 0.713 & 3.498 \\
 ir & {} & {} & {} & 1.46 \\
 \bottomrule
\end{tabular}
\label{tab:results}
\end{table}

\section{Discussion}
The echo state network provides a robust and scalable feature space representation. We transfer this learning representation to a recurrent reinforcement learning agent that learns to target a position directly. It is possible to use the echo state network as a reinforcement learning agent itself, as shown by \citeauthor{ANN_ICANN_2006} \citep{ANN_ICANN_2006}. However, the approach may lead to undesired behaviour in a trading context. Specifically, they use an echo state network to estimate the state-action value function of a temporal difference learning sarsa model \citep{SuttonRichardS1988Ltpb, SuttonBarto2018}. This value function takes the form
\[
  q_\pi(s,a) = \mathbb{E}_\pi \{r_t \lvert s_t, a_t \},
\]
where the expected return $r_t$ depends on the transition to state $s_t$ having taken action $a_t$ under policy $\pi$. Sarsa estimates this value function sequentially as
\begin{equation} \label{eq:sarsa}
  q(s_t, a_t) = (1 - \eta)q(s_t, a_t) + \eta[r_{t+1} + \gamma q(s_{t+1}, a_{t+1})],
\end{equation}
where $0 < \gamma \leq 1$ is a discount factor for multi-step rewards, and $\eta > 0$ is a learning rate. 
Equation \ref{eq:sarsa} shows that the state transition reward is passed back to the starting state. In activities such as maze traversal or board games, being aware of the reward associated with multiple steps or decisions and passing that reward back to the current position is of great value. However, in the context of trading, where the value function $q(s_t, a_t)$ represents the value of a position $s_t$, with the possibility of switching or remaining in the same position denoted by action $a_t$, we will on occasion find that a larger utility is assigned to the wrong state. For example, imagine the current state is $s_t = 0$, that is, the model has no position. Now we observe a large positive price jump leading to large positive reward $r_{t+1} \gg 0$. Value function estimators such as equation \ref{eq:sarsa} would pass the state transition reward $r_{t+1}$ to the initial state $q(s_t=0, a_t=0)$. At the next iteration, with probability $Pr(1-\epsilon)$, $s_{t+1}=0$ as $q(s_{t+1}=0, a_{t+1}=0)$ is the highest value function. However, if the position is zero, the model cannot hope to earn a profit. Even if one excludes the zero state, then there is still the possibility of observing this problem for a reversal strategy with possible states $s = \{-1, 1\}$. Direct reinforcement, as we describe it, does not incur these problems and we have, through transfer learning, improved upon the earlier work in direct reinforcement \citep{Moody1998PerformanceFA, Gold2003} and extended the work of \citeauthor{Borrageiro_RL_FX} \citep{Borrageiro_RL_FX}.

\section{Limitations Of The Work}
\begin{figure}[ht]
 \centering
  \includegraphics[width=.99\columnwidth]{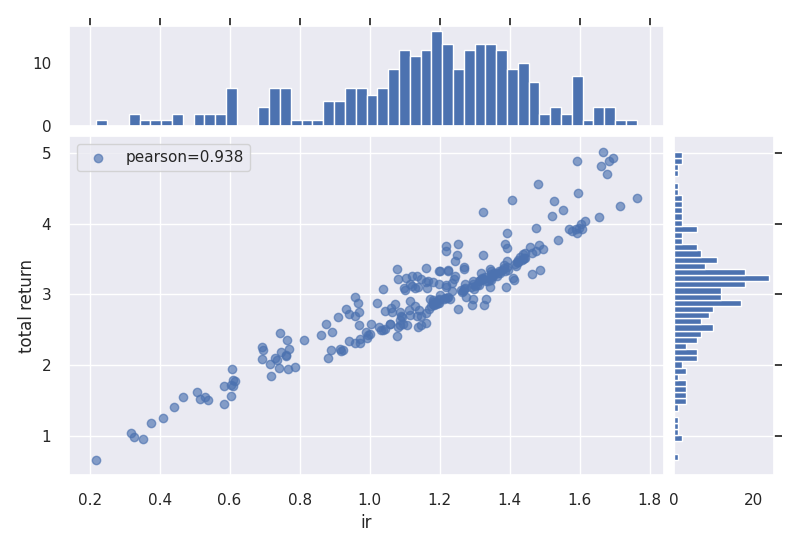}
  \caption{Monte Carlo simulation, information ratio versus total return}
  \label{fig:ir_tr_hist}
\end{figure}

As previously discussed in subsection \ref{sec:dyn_res_feat_space}, the various echo state network parameters are initialised at random. Figure \ref{fig:ir_tr_hist} and table \ref{tab:ir_tr_hist} measure the impact of this randomness on test set performance. We run a Monte Carlo simulation of 250 trials, where the network parameterisation is fixed to $\boldsymbol{\theta} = \{n_{hidden}=100, n_{back}=10 \}$. The information ratios in this set of simulations vary from 0.219 to 1.763, with total returns between 65.4\% and 502.1\%. Whilst this does show evidence of a reasonable variability of returns, table \ref{tab:ir_tr_hist} also shows that 95\% of the mean information ratios vary between 1.1 and 1.2 and 95\% of the mean total returns vary between 289\% and 307\%. Thus the overall picture remains unchanged, that being that the transfer learning crypto agent has successfully learnt how to trade this instrument during the test period. Other than this acceptable sensitivity to weight initialisation, we make no assumptions that would otherwise cause our results to be violated if these assumptions were not met.

\begin{table}[!htb]
\caption{Monte Carlo simulation, information ratio versus total return}
 \centering
 \begin{tabular}{llll}
 \toprule
 {} &    ir & total return \\
 \midrule
 count & 250 &    250 \\
 mean &  1.160 &     2.977 \\
 std  &  0.299 &     0.740 \\
 min  &  0.219 &     0.654 \\
 25\%  &  1.023 &     2.572 \\
 50\%  &  1.199 &     3.076 \\
 75\%  &  1.355 &     3.352 \\
 max  &  1.763 &     5.021 \\
 se(mean) &  0.019 &     0.047 \\
 lb(mean) &  1.123 &     2.886 \\
 ub(mean) &  1.197 &     3.069 \\
 \bottomrule
\end{tabular}
\label{tab:ir_tr_hist}
\end{table}

\section{Conclusion}
We demonstrate an application of online transfer learning as a digital assets trading agent. This agent uses a powerful feature space representation in the form of an echo state network, the output of which is made available to a direct, recurrent reinforcement learning agent. The agent learns to trade the XBTUSD (Bitcoin versus US Dollars) perpetual swap derivatives contract on BitMEX. It learns to trade intraday on five minutely sampled data, avoids excessive over-trading, captures a funding profit and is also able to predict the direction of the market. Overall, our crypto agent realises a total return of 350\%, net of transaction costs, over roughly five years, 71\% of which is down to funding profit. The annualised information ratio that it achieves is 1.46.

\typeout{}
\EOD


\bibliographystyle{IEEEtranN} 
\bibliography{references}

\begin{thebibliography}{44}
\providecommand{\natexlab}[1]{#1}
\providecommand{\url}[1]{#1}
\csname url@samestyle\endcsname
\providecommand{\newblock}{\relax}
\providecommand{\bibinfo}[2]{#2}
\providecommand{\BIBentrySTDinterwordspacing}{\spaceskip=0pt\relax}
\providecommand{\BIBentryALTinterwordstretchfactor}{4}
\providecommand{\BIBentryALTinterwordspacing}{\spaceskip=\fontdimen2\font plus
\BIBentryALTinterwordstretchfactor\fontdimen3\font minus
  \fontdimen4\font\relax}
\providecommand{\BIBforeignlanguage}[2]{{%
\expandafter\ifx\csname l@#1\endcsname\relax
\typeout{** WARNING: IEEEtranN.bst: No hyphenation pattern has been}%
\typeout{** loaded for the language `#1'. Using the pattern for}%
\typeout{** the default language instead.}%
\else
\language=\csname l@#1\endcsname
\fi
#2}}
\providecommand{\BIBdecl}{\relax}
\BIBdecl

\bibitem[Shumway(2011)]{ShumwayRobertH2011Tsaa}
R.~H. Shumway, \emph{Time series analysis and its applications : with R
  examples}, 3rd~ed., ser. Springer texts in statistics.\hskip 1em plus 0.5em
  minus 0.4em\relax Springer, 2011.

\bibitem[Granger and Newbold(1974)]{GrangerNewbold1974}
C.~W.~J. Granger and P.~Newbold, ``Spurious regressions in econometrics,''
  \emph{Journal of econometrics}, vol.~2, pp. 111--120, 1974.

\bibitem[Borrageiro et~al.(2022)Borrageiro, Firoozye, and
  Barucca]{Borrageiro_RL_FX}
G.~Borrageiro, N.~Firoozye, and P.~Barucca, ``Reinforcement learning for
  systematic fx trading,'' \emph{IEEE Access}, vol.~10, pp. 5024--5036, 2022.

\bibitem[Yang et~al.(2020)Yang, Zhang, Dai, and
  Pan]{yang_transfer_learning_2020}
Q.~Yang, Y.~Zhang, W.~Dai, and S.~J. Pan, \emph{Transfer Learning}.\hskip 1em
  plus 0.5em minus 0.4em\relax Cambridge University Press, 2020.

\bibitem[Sutton and Barto(2018)]{SuttonBarto2018}
R.~S. Sutton and A.~G. Barto, \emph{Reinforcement Learning: An
  Introduction}.\hskip 1em plus 0.5em minus 0.4em\relax Cambridge, MA, USA: A
  Bradford Book, 2018.

\bibitem[Jaeger(2002)]{Jaeger2002AdaptiveNS}
H.~Jaeger, ``Adaptive nonlinear system identification with echo state
  networks,'' in \emph{NIPS}, 2002.

\bibitem[Moody et~al.(1998)Moody, Wu, Liao, and
  Saffell]{Moody1998PerformanceFA}
J.~Moody, L.~Wu, Y.~Liao, and M.~Saffell, ``Performance functions and
  reinforcement learning for trading systems and portfolios,'' \emph{Journal of
  Forecasting}, vol.~17, pp. 441--470, 1998.

\bibitem[Dacorogna et~al.(2001)Dacorogna, Gencay, Altunç, Pictet, and
  Olsen]{Dacorogna2001AnIT}
M.~M. Dacorogna, R.~Gencay, S.~Altunç, O.~V. Pictet, and R.~Olsen, \emph{An
  Introduction to High-Frequency Finance}.\hskip 1em plus 0.5em minus
  0.4em\relax Academic Press, 2001.

\bibitem[Zhao et~al.(2014)Zhao, Hoi, Wang, and Li]{Zhao_OTL_2014}
P.~Zhao, S.~C.~H. Hoi, J.~Wang, and B.~Li, ``Online transfer learning,''
  \emph{Artificial intelligence}, vol. 216, pp. 76--102, 2014.

\bibitem[Salvalaio and de~Oliveira~Ramos(2019)]{Bruno_OTL_2019}
B.~K. Salvalaio and G.~de~Oliveira~Ramos, ``Self-adaptive appearance-based
  eye-tracking with online transfer learning,'' in \emph{2019 8th Brazilian
  Conference on Intelligent Systems (BRACIS)}.\hskip 1em plus 0.5em minus
  0.4em\relax IEEE, 2019, pp. 383--388.

\bibitem[Wang et~al.(2020)Wang, Wang, and Zeng]{Wang_OTL_2020}
X.~Wang, X.~Wang, and Z.~Zeng, ``A novel weight update rule of online transfer
  learning,'' in \emph{2020 12th International Conference on Advanced
  Computational Intelligence (ICACI)}.\hskip 1em plus 0.5em minus 0.4em\relax
  IEEE, 2020, pp. 349--355.

\bibitem[Pan and Yang(2010)]{Pan2010ASO}
S.~J. Pan and Q.~Yang, ``A survey on transfer learning,'' \emph{IEEE
  Transactions on Knowledge and Data Engineering}, vol.~22, pp. 1345--1359,
  2010.

\bibitem[Lukosevicius et~al.(2012)Lukosevicius, Jaeger, and
  Schrauwen]{Lukosevicius2012ReservoirCT}
M.~Lukosevicius, H.~Jaeger, and B.~Schrauwen, ``Reservoir computing trends,''
  \emph{KI - K{\"u}nstliche Intelligenz}, vol.~26, pp. 365--371, 2012.

\bibitem[Lukosevicius and Jaeger(2009)]{Lukosevicius2009ReservoirCA}
M.~Lukosevicius and H.~Jaeger, ``Reservoir computing approaches to recurrent
  neural network training,'' \emph{Comput. Sci. Rev.}, vol.~3, pp. 127--149,
  2009.

\bibitem[Murray(2019)]{Murray2019LocalOL}
J.~M. Murray, ``Local online learning in recurrent networks with random
  feedback,'' \emph{eLife}, vol.~8, 2019.

\bibitem[Maass and Markram(2004)]{Maass2004OnTC}
W.~Maass and H.~Markram, ``On the computational power of circuits of spiking
  neurons,'' \emph{J. Comput. Syst. Sci.}, vol.~69, pp. 593--616, 2004.

\bibitem[Samadi et~al.(2017)Samadi, Lillicrap, and Tweed]{Samadi2017DeepLW}
A.~Samadi, T.~P. Lillicrap, and D.~B. Tweed, ``Deep learning with dynamic
  spiking neurons and fixed feedback weights,'' \emph{Neural Computation},
  vol.~29, pp. 578--602, 2017.

\bibitem[Szita et~al.(2006)Szita, Gyenes, and L{\H{o}}rincz]{ANN_ICANN_2006}
I.~Szita, V.~Gyenes, and A.~L{\H{o}}rincz, ``Reinforcement learning with echo
  state networks,'' in \emph{Artificial Neural Networks -- ICANN 2006}.\hskip
  1em plus 0.5em minus 0.4em\relax Berlin, Heidelberg: Springer Berlin
  Heidelberg, 2006, pp. 830--839.

\bibitem[Gordon(2000)]{Gordon2000ReinforcementLW}
G.~J. Gordon, ``Reinforcement learning with function approximation converges to
  a region,'' in \emph{NIPS}, 2000.

\bibitem[Shi et~al.(2017)Shi, Liu, and Wei]{Shi2017EchoSN}
G.~Shi, D.~Liu, and Q.~Wei, ``Echo state network-based q-learning method for
  optimal battery control of offices combined with renewable energy,''
  \emph{Iet Control Theory and Applications}, vol.~11, pp. 915--922, 2017.

\bibitem[Chen et~al.(2021)Chen, Jin, and Song]{CHEN2021}
Q.~Chen, Y.~Jin, and Y.~Song, ``Fault-tolerant adaptive tracking control of
  euler-lagrange systems – an echo state network approach driven by
  reinforcement learning,'' \emph{Neurocomputing}, 2021.

\bibitem[Luko{\v{s}}evi{\v{c}}ius(2012)]{Luko2012}
M.~Luko{\v{s}}evi{\v{c}}ius, \emph{A Practical Guide to Applying Echo State
  Networks}.\hskip 1em plus 0.5em minus 0.4em\relax Berlin, Heidelberg:
  Springer Berlin Heidelberg, 2012, pp. 659--686.

\bibitem[Bishop(1994)]{Bishop_1994}
C.~M. Bishop, ``Neural networks and their applications,'' \emph{Review of
  Scientific Instruments}, vol.~65, no.~6, pp. 1803--1832, 1994.

\bibitem[Williams(1992{\natexlab{a}})]{Williams1992}
R.~J. Williams, ``Simple statistical gradient-following algorithms for
  connectionist reinforcement learning,'' \emph{Machine Learning}, vol.~8, 5
  1992.

\bibitem[Sugiyama(2015)]{Sugiyama2015}
M.~Sugiyama, \emph{Statistical Reinforcement Learning}, 1st~ed.\hskip 1em plus
  0.5em minus 0.4em\relax CRC Press, 2015.

\bibitem[Sharpe(1966)]{Sharpe1966}
W.~F. Sharpe, ``Mutual fund performance,'' \emph{The Journal of Business},
  vol.~39, 1 1966.

\bibitem[Rumelhart et~al.(1985)Rumelhart, Hinton, and
  Williams]{Rumelhart1986LearningIR}
D.~E. Rumelhart, G.~E. Hinton, and R.~J. Williams, ``Learning internal
  representations by error propagation,'' California Univ San Diego La Jolla
  Inst for Cognitive Science, Tech. Rep., 1985.

\bibitem[Werbos(1990)]{Werbos_bptt}
P.~Werbos, ``Backpropagation through time: what it does and how to do it,''
  \emph{Proceedings of the IEEE}, vol.~78, no.~10, pp. 1550--1560, 1990.

\bibitem[Williams and Zipser(1989)]{RNN1989}
R.~J. Williams and D.~Zipser, ``A learning algorithm for continually running
  fully recurrent neural networks,'' \emph{Neural computation}, vol.~1, pp.
  270--280, 1989.

\bibitem[Gold(2003)]{Gold2003}
C.~Gold, ``Fx trading via recurrent reinforcement learning,'' in
  \emph{IEEE}.\hskip 1em plus 0.5em minus 0.4em\relax IEEE, 2003.

\bibitem[Jiang and Liang(2017)]{crypto_portfolio_2017}
Z.~Jiang and J.~Liang, ``Cryptocurrency portfolio management with deep
  reinforcement learning,'' in \emph{2017 Intelligent Systems Conference
  (IntelliSys)}, 2017, pp. 905--913.

\bibitem[Lee et~al.(2018)Lee, Ulkuatam, Beling, and Scherer]{lee2018}
K.~Lee, S.~Ulkuatam, P.~Beling, and W.~Scherer, ``Generating synthetic bitcoin
  transactions and predicting market price movement via inverse reinforcement
  learning and agent-based modeling,'' \emph{Journal of Artificial Societies
  and Social Simulation}, vol.~21, no.~3, p.~5, 2018.

\bibitem[Russell(1998)]{inverse_RL}
S.~Russell, ``Learning agents for uncertain environments,'' in
  \emph{Proceedings of the Eleventh Annual Conference on Computational Learning
  Theory}, ser. COLT' 98.\hskip 1em plus 0.5em minus 0.4em\relax New York, NY,
  USA: Association for Computing Machinery, 1998, p. 101–103.

\bibitem[Lucarelli and Borrotti(2019)]{deep_RL_crypto_2019}
G.~Lucarelli and M.~Borrotti, ``A deep reinforcement learning approach for
  automated cryptocurrency trading,'' in \emph{Artificial Intelligence
  Applications and Innovations}.\hskip 1em plus 0.5em minus 0.4em\relax Cham:
  Springer International Publishing, 2019, pp. 247--258.

\bibitem[Mnih et~al.(2013)Mnih, Kavukcuoglu, Silver, Graves, Antonoglou,
  Wierstra, and Riedmiller]{mnih2013playing}
V.~Mnih, K.~Kavukcuoglu, D.~Silver, A.~Graves, I.~Antonoglou, D.~Wierstra, and
  M.~Riedmiller, ``Playing atari with deep reinforcement learning,'' 2013.

\bibitem[Wang et~al.(2016)Wang, Schaul, Hessel, Van~Hasselt, Lanctot, and
  De~Freitas]{dueling_net_2016}
Z.~Wang, T.~Schaul, M.~Hessel, H.~Van~Hasselt, M.~Lanctot, and N.~De~Freitas,
  ``Dueling network architectures for deep reinforcement learning,'' in
  \emph{Proceedings of the 33rd International Conference on International
  Conference on Machine Learning - Volume 48}, ser. ICML'16.\hskip 1em plus
  0.5em minus 0.4em\relax JMLR.org, 2016, p. 1995–2003.

\bibitem[Zhang et~al.(2022)Zhang, Zhao, Wu, Li, Huang, and
  Tan]{portfolio_deep_RL_2022}
Y.~Zhang, P.~Zhao, Q.~Wu, B.~Li, J.~Huang, and M.~Tan, ``Cost-sensitive
  portfolio selection via deep reinforcement learning,'' \emph{IEEE
  Transactions on Knowledge and Data Engineering}, vol.~34, no.~1, pp.
  236--248, 2022.

\bibitem[Yildiz et~al.(2012)Yildiz, Jaeger, and Kiebel]{Yildiz2012RevisitingTE}
I.~B. Yildiz, H.~Jaeger, and S.~J. Kiebel, ``Re-visiting the echo state
  property,'' \emph{Neural networks : the official journal of the International
  Neural Network Society}, vol.~35, pp. 1--9, 2012.

\bibitem[Jaeger(2017)]{Jaeger2017UsingCT}
H.~Jaeger, ``Using conceptors to manage neural long-term memories for temporal
  patterns,'' \emph{Journal of Machine Learning Research}, vol.~18, pp.
  13:1--13:43, 2017.

\bibitem[Sharpe(2007)]{Sharpe2007}
W.~F. Sharpe, ``Expected utility asset allocation,'' \emph{Financial Analysts
  Journal}, vol.~63, no.~5, pp. 18--30, 2007.

\bibitem[Kahn(2019)]{grinold2019advances}
R.~C. G. R.~N. Kahn, \emph{Advances in Active Portfolio Management: New
  Developments in Quantitative Investing}.\hskip 1em plus 0.5em minus
  0.4em\relax McGraw-Hill Education, 2019.

\bibitem[Williams(1992{\natexlab{b}})]{Williams_rnn_ekf}
R.~J. Williams, ``Training recurrent networks using the extended kalman
  filter,'' in \emph{[Proceedings 1992] IJCNN International Joint Conference on
  Neural Networks}, vol.~4, 1992, pp. 241--246 vol.4.

\bibitem[Haykin(2001)]{Haykin2001}
S.~Haykin, \emph{Kalman filtering and neural networks}.\hskip 1em plus 0.5em
  minus 0.4em\relax Wiley, 2001.

\bibitem[Sutton(1988)]{SuttonRichardS1988Ltpb}
R.~S. Sutton, ``\BIBforeignlanguage{eng}{Learning to predict by the methods of
  temporal differences},'' \emph{\BIBforeignlanguage{eng}{Machine learning}},
  vol.~3, no.~1, pp. 9--44, 1988.

\end{thebibliography}
\vspace{-11mm}

\begin{IEEEbiography}[{\includegraphics[width=1in,height=1.25in,clip,keepaspectratio]{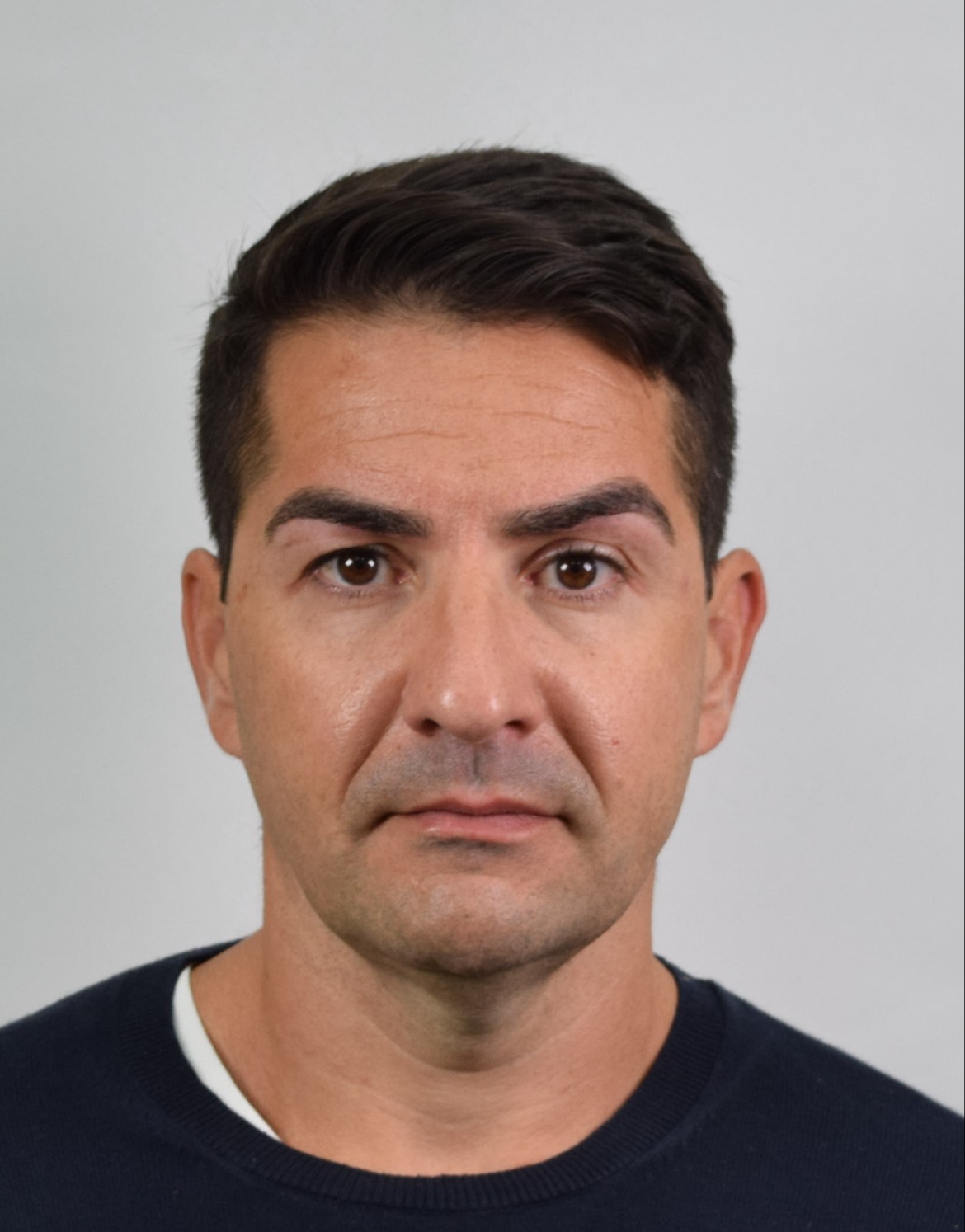}}]{Gabriel Borrageiro} is a PhD student at the University College London, in the Computer Science department and part of the Financial Computing Group. His research interests include online learning, reinforcement learning and neural networks. He obtained his executive MBA from Cass Business School, City University, London, and a computer science diploma from Damelin College, South Africa. Gabriel is employed as a quantitative researcher at BlueCrest Capital.
\end{IEEEbiography}
\vspace{-13mm}

\begin{IEEEbiography}[{\includegraphics[width=1in,height=1.25in,clip,keepaspectratio]{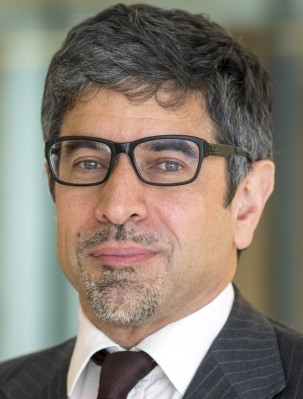}}]{Nick Firoozye} is currently Honorary Reader, Computational Finance, in the Computer Science department at University College London and also part of the Financial Computing Group. He obtained his PhD and MS in mathematics at New York University and a BS in mathematics at Harvey Mudd College. He also works for Exos Bank in the systematic rates trading business.
\end{IEEEbiography}
\vspace{-13mm}

\begin{IEEEbiography}[{\includegraphics[width=1in,height=1.25in,clip,keepaspectratio]{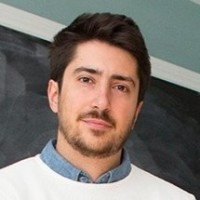}}]{Paolo Barucca} is a lecturer at University College London, Computer Science department and also part of the Financial Computing Group. He is also editor in chief of the science dissemination project, La Scienza Coatta and scientific officer of the Blockchain Education Network. Paolo received his PhD in theoretical and mathematical physics from Sapienza Universita di Roma.
\end{IEEEbiography}

\end{document}